\documentclass{article}

\usepackage[nonatbib, final]{neurips_2021}
\usepackage[sort, compress, square, numbers]{natbib}
\usepackage[utf8]{inputenc} % allow utf-8 input
\usepackage[T1]{fontenc}    % use 8-bit T1 fonts
\usepackage{hyperref}       % hyperlinks
\usepackage{url}            % simple URL typesetting
\usepackage{booktabs}       % professional-quality tables
\usepackage{amsfonts}       % blackboard math symbols
\usepackage{nicefrac}       % compact symbols for 1/2, etc.
\usepackage{microtype}      % microtypography
\usepackage{xcolor}         % colors
\usepackage{amsmath}
\usepackage{amssymb}
\usepackage{url}
\usepackage{makecell}
\usepackage{multirow}
\usepackage{booktabs}
\usepackage{array}

\usepackage[utf8]{inputenc}

\usepackage{microtype}
\usepackage{kotex}

\newcommand{\thickhline}{%
    \noalign {\ifnum 0=`}\fi \hrule height 1pt
    \futurelet \reserved@a \@xhline
}

\usepackage{comment}
\usepackage{todonotes}
\usepackage{xcolor}
\title{Automatic Knowledge Augmentation for Generative Commonsense Reasoning}

\author{Jaehyung Seo\\
  Korea University \\ \texttt{seojae777@korea.ac.kr} \\
\And
Chanjun Park \\
Korea University \\
\texttt{bcj1210@korea.ac.kr} \\

\And
Sugyeong Eo \\
Korea University \\
\texttt{djtnrud@korea.ac.kr} \\

\And
Hyeonseok Moon \\
Korea University \\
\texttt{glee889@korea.ac.kr} \\

\And
Heuiseok Lim$^{\dagger}$ \\
Korea University \\
\texttt{limhseok@korea.ac.kr} \\

}

\begin{document}

\maketitle

\begin{abstract}
Generative commonsense reasoning is the capability of a language model to generate a sentence with a given concept-set that is based on commonsense knowledge. However, generative language models still struggle to provide outputs, and the training set does not contain patterns that are sufficient for generative commonsense reasoning. In this paper, we propose a data-centric method that uses automatic knowledge augmentation to extend commonsense knowledge using a machine knowledge generator. This method can generate semi-golden sentences that improve the generative commonsense reasoning of a language model without architecture modifications. Furthermore, this approach is a model-agnostic method and does not require human effort for data construction.
\end{abstract}

\section{Introduction}
CommonGen \citep{lin2020commongen} is a challenge in which the task is to generate a sentence that makes sense with an input concept-set by leveraging generative commonsense reasoning. However, it is difficult to derive much commonsense knowledge for the given concept-sets, such as compositional generalization and relational reasoning, from the 67,389 sentences in the CommonGen training set \citep{lake2018generalization, keysers2019measuring}. Additionally, most reference sentences consist of superficial scene descriptions and do not represent any specific relational knowledge between the concepts \citep{lin2020commongen}. 

To address these issues, we propose a data-centric approach that augments prior knowledge with an additional pre-training session. Our method uses a trained generator that produces semi-golden sentences as patterns to guide generative commonsense reasoning \citep{NEURIPS2020_6b493230,yang2020g}. As the generator, we utilize BART \citep{lewis2019bart}, which has the highest average percentage of input concepts that exist in the output sentences in the CommonGen task ($97.35\%$). The generator learns to reconstruct sentences using concepts (e.g., verbs and nouns) extracted from sentences in Wikipedia articles, which have high coverage. Semi-golden sentences with concrete descriptions about the input concept-set or concept-pair set are used in an additional pre-training session to enhance the prior knowledge of the generative language model.

% 더 나아가, autoregressive language model인 GPT2와 sequence-to-sequence lm인 T5에 대해 실험하여, 우리 방식의 강력한 일반화 가능성과 model-agnostic함을 증명한다. 
Automatic knowledge generation substantially reduces human effort and ensures the quality of machine-generated data using the results of performance improvements. It can also extend the relational information of given concept-sets, including commonsense knowledge for the intended input query. Moreover, the improvements in performance in auto-regressive models ({\em{i.e.,}} a decoder only) \citep{radford2019language} and sequence-to-sequence language model \citep{raffel2020exploring} ({\em{i.e.,}} an encoder--decoder), which have different architectures, demonstrate the strong generalization and model-agnostic nature of this approach. 

\begin{table}
  \caption{Experimental results and statistics for each data augmentation used in pre-training. ``Pair'' and ``Set'' respectively indicate to the semi-golden sentences for concept-pairs and concept-sets in the CommonGen training set; ``\# sentences'' and ``\# concept-sets'' respectively refer to the number of semi-golden sentences and number of concepts with respect to the size of the concept-set.}
  \label{sample-table}
  \centering
  \resizebox{\linewidth}{!}{\begin{tabular}{lccccccc}
    \toprule 
    & \multicolumn{5}{c}{\textbf{Results}} & \multicolumn{2}{c}{\textbf{Augmentation}}
    \\\cmidrule(lr){2-6}\cmidrule(lr){7-8}
           & ROUGE-L & BLEU4 & METEOR & CIDEr & SPICE    & \# sentences & \# concept-sets
    \\\midrule
    \textbf{GPT2}    & 39.28 & 21.10 & 26.20 & 12.15 & 25.90 & - & - \\
    \textbf{GPT2+Pair} & \textbf{49.70} & 23.10 & 26.90 & 12.62 & 27.20 & 59,125 & 2(59,125) \\
    \textbf{GPT2+Set} & 49.20 & 22.30 & 26.80 & 12.60 & 27.10 & 32,471 & 3(24,891), 4(4,206), 5(3,374)\\
    \textbf{GPT2+Pair+Set} & 49.50 & \textbf{23.50} & \textbf{27.20} & \textbf{12.95} & \textbf{27.60} & 91,596 &2(59,125), 3(24,891), 4(4,206) 5(3,374)\\
    \midrule
    \textbf{T5}    & 49.27 & 28.60 & 30.10 & 14.96 & 31.60 & - & - \\
    \textbf{T5+Pair} & 56.90 & 33.70 & 33.00 & 17.72 & 33.80 & 59,125 & 2(59,125) \\
    \textbf{T5+Set} & \textbf{57.20} & \textbf{34.40} & \textbf{33.00} & \textbf{17.97} & \textbf{33.90} & 32,471 & 3(24,891), 4(4,206), 5(3,374)\\
    \textbf{T5+Pair+Set} & 56.80 & 33.40 & 32.90 & 17.76 & 33.80 & 91,596 &2(59,125), 3(24,891), 4(4,206) 5(3,374)
    \\\bottomrule
    \end{tabular}}
\end{table}

\section{Data Construction Process and Results}

In this section, we present the process of data construction and experiments using a generator.

\textbf{First}, we extracted 154,892 sentences from a Wikipedia data dump (2021-03-01) that match more than two input concepts via ElasticSearch \citep{divya2013} and BM25 \citep{svore2009machine, chen2017reading}. From the extracted sentences $y$, we defined concept-set $C$, which includes a verb or noun through part-of-speech (POS) tagging. \textbf{Second}, the generator was trained with the reconstructed data using $C = y$ and produced a semi-golden sentence $y'$ for an input concept-set $C'$. \textbf{Third}, 59,125 concept-pairs and 32,471 unique concept-sets (of size 3 to 5) were extracted from the CommonGen training set and the generator used them as an input sequence to create a semi-golden sentence. \textbf{Fourth}, extracted concept-sets and generated semi-golden sentences were used as pre-training data in the form of $C' = y'$ for two generative language models, as presented in Table 1. We conducted a case study to demonstrate the impact of our method with GPT2 \citep{radford2019language} and T5 \citep{raffel2020exploring}. To evaluate the performance, we used the n-gram overlapping metrics BLEU \citep{papineni2002bleu}, ROUGE \citep{lin2004rouge}, and METEOR   \citep{banerjee2005meteor} as well as the concept matching metrics CIDEr \citep{vedantam2015cider} and SPICE \citep{anderson2016spice}.  

As shown in Table 1, all data augmentation cases lead to performance improvements with respect to all evaluation metrics. The results also indicate that there is a difference in performance depending on the combination of input concepts and target models. Augmented concept-pairs include concrete relational knowledge between two concepts and concept-sets have commonsense knowledge for generating sentences by compositionality. GPT2 obtains the best performance improvement when learning both types of machine-generated knowledge through the augmented dataset. In contrast, T5 yields no significant performance difference depending on the form of the augmented dataset, although learning commonsense knowledge with compositionality is slightly more effective. This result shows that there is a variance in the degree of improvement in prior knowledge in a specific area when the model architecture and dataset used for pre-training each language model differ. This experimental result demonstrates that our data-centric approach generates better semi-golden sentences for use as an augmented dataset. It is also possible to produce a machine-generated dataset in the intended format to compensate for the insufficient prior knowledge of each model. Furthermore, this approach is a model-agnostic method and has a positive impact on both auto-regressive and sequence-to-sequence architectures.

\section{Conclusion} 
We demonstrated how to use natural language generation for data-centric research to reduce the cost of leveraging information retrieval and building a human-annotated dataset. Automatic knowledge augmentation for generative commonsense reasoning is also not limited to a particular model or dataset, but can be applied in different forms as needed. In future work, this method will be highly valuable because it can be extended to other downstream tasks that require commonsense knowledge as supporting evidence.

\section*{Acknowledgment}
This research was supported by the MSIT(Ministry of Science and ICT), Korea, under the ITRC(Information Technology Research Center) support program(IITP-2018-0-01405) supervised by the IITP(Institute for Information \& Communications Technology Planning \& Evaluation) and IITP grant funded by the Korea government(MSIT) (No. 2020-0-00368, A Neural-Symbolic Model for Knowledge Acquisition and Inference Techniques) and Basic Science Research Program through the National Research Foundation of Korea(NRF) funded by the Ministry of Education(NRF-2021R1A6A1A03045425). Heuiseok Lim$^\dagger$ is a corresponding author.

\bibliography{ref}
\bibliographystyle{acl_natbib}

\end{document}